\documentclass[letterpaper]{article} 
\usepackage{aaai19}  
\usepackage{times}  
\usepackage{helvet}  
\usepackage{courier}  
\usepackage{url}  
\usepackage{graphicx}  
\usepackage{float} 
\usepackage{wasysym}

\begin{document}
%
\title{How Do Fairness Definitions Fare? \\
Examining Public Attitudes Towards Algorithmic Definitions of Fairness$^*$ }
\author{Nripsuta Ani Saxena$^{**}$ \\ University of Southern California \And Karen Huang\\ Harvard University \And Evan DeFilippis\\ Harvard University \AND Goran Radanovic\\ Harvard University \And David C. Parkes\\ Harvard University \And Yang Liu$^{**}$ \\ University of California, Santa Cruz}

\maketitle
\begin{abstract}
  What is the best way to define algorithmic fairness?  While many
  definitions of fairness have been proposed in the computer science
  literature, there is no clear agreement over a particular
  definition. In this work, we investigate ordinary people's
  perceptions of three of these fairness definitions. Across two
  online experiments, we test which definitions people perceive to be
  the fairest in the context of loan decisions, and whether fairness
  perceptions change with the addition of sensitive information (i.e.,
  race of the loan applicants). Overall, one definition (calibrated
  fairness) tends to be more preferred than the others, and the
  results also provide support for the principle of affirmative
  action.
\end{abstract}

\let\thefootnote\relax\footnotetext{$^*$ Preconference version: Not final.}

\let\thefootnote\relax\footnotetext{$^{**}$ Correspondence should be directed to: nsaxena@usc.edu and yangliu@ucsc.edu.}

\section{Introduction}
\noindent Algorithms are increasingly being used in high-impact domains of decision-making, such as loans, hiring, bail, and university admissions, with wide-ranging societal implications. However, issues have arisen regarding the fairness of these algorithmic decisions. For example, the risk assessment software, COMPAS, used by judicial systems in many states, predicts a score indicating the likelihood of a defendant committing a crime if given bail. ProPublica analyzed recidivism predictions from COMPAS for criminal defendants, and looked at false positive rates and false negative rates for defendants of different races. It argued that the tool is biased against black defendants \cite{angwin2016machine}. Equivant (formerly called Northpointe), the company that developed the COMPAS tool, on the other hand, focused on positive predictive value, which is similar for whites and blacks \cite{dieterich2016compas}. That is, by some measures of fairness, the tool was found to be biased against blacks; meanwhile by other measures, it was not. Which measures are fair?

The above scenario is not a rare case. Given the increasing pervasiveness of automated decision-making systems, there's a growing concern among both computer scientists and the public for how to ensure algorithms are fair. While several definitions of fairness have recently been proposed in the computer science literature, there's a lack of agreement among researchers about which definition is the most appropriate \cite{gajane2017formalizing}. It is very unlikely that one definition of fairness will be sufficient. This is supported also by recent impossibility results that show some fairness definitions cannot coexist \cite{kleinberg2016inherent}. Since the public is affected by these algorithmic systems, it is important to investigate public views of algorithmic fairness  \cite{lee2017algorithmic,lee2017human,lee2018understanding,binns2018s,woodruff2018qualitative}.

While substantial research has been done in moral psychology to understand people's perceptions of fairness (e.g, \citeauthor{yaari1984dividing} (1984), \cite{bazerman1995perceptions} (1995)), relatively little work has been done to understand how the general public views fairness criteria in algorithmic decision making: \citeauthor{pierson2017gender} (2017) investigated how two different factors influence views on algorithmic fairness, \citeauthor{plane2017exploring}  (2017) explored human perceptions of discrimination in targeted online advertising, \citeauthor{grgic2018beyond} (2018), \citeauthor{grgic2018human} (2018) studied human perceptions of features used in algorithmic decision making, \citeauthor{binns2018s} (2018) examined people's perception of justice  in algorithmic decision making under different explanation styles. In contrast to this work, our goal is to understand how people perceive the fairness definitions proposed in the recent computer science literature, that is, the outcomes allowed by these definitions.

By testing people's perception of different fairness definitions, we hope to spur more work on understanding definitions of fairness that are appropriate for particular contexts. In line with recent work examining public attitudes of the ethical programming of machines  \cite{bonnefon2016social,awad2018moral}, we suggest that these public attitudes serve as a useful and important input in a conversation between technologists and ethicists. These findings can help technologists to develop decision-making algorithms with fairness principles aligned with those of the general public, to make sure that designs are sensitive to the prevailing notions of fairness in society. Crowdsourcing can also be used to understand how preferences vary across geographies and cultures.

\section{Definitions of Fairness}

Broadly, we investigate a concept of fairness known as \textit{distributive justice}, or fairness regarding the outcomes \cite{adams1963towards,adams1965inequity}. However, which characteristics regarding the individual should be relevant and which should be irrelevant to fairness? We instantiate our study via investigating two characteristics: task-specific similarity (loan repayment rate) and a sensitive attribute (race), and collect data on attitudes toward the relevancy of these characteristics. In principle, fairness is the absence of any bias based on an individual's inherent or acquired characteristics that are irrelevant in the particular context of decision-making \cite{chouldechova2017fair}. In many contexts, these inherent characteristics (referred to as `sensitive attributes' or `protected attributes' in the computer science literature), are gender, religion, race, skin color, age, or national origin.

We restrict our emphasis to three fairness definitions from the computer science literature. We choose to test these three definitions because these definitions can be easily operationalized as distinct decisions in the context of loan scenarios that are easily understandable by ordinary people. In our experiments, we map these definitions (or constrained versions of the definitions) to distinct loan allocation choices, and test people's judgments of these choices. We summarize the three fairness definitions as follows:

\noindent
\textbf{Treating similar individuals similarly.} \citeauthor{dwork2012fairness} (2012) formulate fairness as treating similar individuals (with respect to certain attributes) similarly in receiving a favorable decision, where the similarity of any two individuals is determined on the basis of a similarity distance metric, specific to the task at hand, and that ideally represents a notion of ground truth in regard to the decision context. Given this similarity metric, an algorithm would be fair if its decisions satisfied the Lipschitz condition (a continuity and similarity measure) defined with respect to the metric. In our loan allocation scenario, individuals with similar repayment rates should receive similar amounts of money.

\noindent
\textbf{Never favor a worse individual over a better one.} In the context of online learning, \citeauthor{joseph2016fairness} (2016) define fairness, in a setting where a single individual is to be selected for a favorable decision, as always choosing a better individual (with higher expected value of some measure of inherent quality) with a probability greater than or equal to the probability of choosing a worse individual. This definition promotes meritocracy with respect to the candidate's inherent quality. \citeauthor{joseph2016fairness} (2016) apply this definition of fairness to the setting of contextual bandits, a classical sequential decision-making process, by utilizing the expected reward to determine the quality of an action (an arm as in the bandit setting). Each arm represents a different subpopulation, and each subpopulation may have its own function that maps decision context to expected payoff. In our loan allocation scenario, an individual with a higher repayment rate should obtain at least as much money as her peer.

\noindent
\textbf{Calibrated fairness.}
The third definition, that we refer to as `calibrated fairness', is formulated by \citeauthor{liu2017calibrated} (2017) in the setting of sequential decision-making$^{[1]}$. \footnote{$^{[1]}$Note that \citeauthor{kleinberg2016inherent} (2016), \citeauthor{chouldechova2017fair} (2017) define `calibration' in a different way, that includes the notion of a sensitive attribute.} Calibrated fairness selects individuals in proportion to their merit. In a multi-armed bandit setting, this means that an arm would be pulled with a probability that its pull would result the largest reward if all the arms are pulled. When the merit is known (underlying true quality), calibrated fairness implies the meritocratic fairness of  \citeauthor{joseph2016fairness} (2016). Furthermore, as argued by \citeauthor{liu2017calibrated} (2017), calibrated fairness implies \citeauthor{dwork2012fairness} (2016) for a suitably chosen similarity metric. In our loan allocation scenario, we interpret calibrated fairness as requiring that two individuals with repayment rates $r_1$ and $r_2$, respectively, should  obtain $r_1/(r_1 + r_2)$ and $r_2/(r_1 + r_2)$ amount of money, respectively$^{[2]}$. \footnote{$^{[2]}$This is a slightly different version of the formal definition in \citeauthor{liu2017calibrated} (2017), which would take the ratio in proportion to the rate at which one individual repays while the other does not, but we feel a more intuitive way to capture the idea of  calibrated fairness in our setting.)}

\section{Overview of Present Research}
In the present research, we ask: when do people endorse one fairness definition over another?

First, we want to understand how support for the three definitions of fairness depends on variation in the similarity of the target individuals.  The three definitions differ in how this comparison between task-specific metrics should matter. 

We are also interested to understand how information about the race of the two target individuals influences these fairness perceptions. Direct discrimination is the phenomenon of discriminating against an individual simply because of their membership, or perceived membership, in certain protected (or sensitive) attributes, such as age, disability, religion, gender, and race \cite{ellis2012eu}. All three definitions agree that, conditioned on the relevant task-specific metric, an attribute such as race should not be relevant to decision-making.$^{[3]}$\footnote{$^{[3]}$Here, we assume that the treating similar individuals similarly definition~\cite{dwork2012fairness} 
does not use race as a relevant dimension for judging individual similarity.}
%
%
Information about race may matter, however, since people may consider race to be an important factor for distributive justice. For example, in decisions promoting affirmative action, people may believe that considering race is important in order to address historical inequities. If that is the case, then definitions of algorithmic fairness may need to take into account such sensitive attributes.

Across two online experiments, we investigate how people perceive algorithmic fairness in the context of loans, which is a setting with a divisible good to allocate. We employ a scenario where a loan officer must decide how to allocate a limited amount of loan money to two individuals. In Study 1, we test how the individuals' task-specific similarity (i.e., loan repayment rates) influences perceptions of fairness, in the absence of information about race. In Study 2, we test how the individuals' race may, along with their loan repayment rates, influence perceptions of fairness. For the purpose of the study, we need to interpret these fairness definitions, which are formalized for choosing a single individual for a favorable decision (or assigning an indivisible good) to this setting where the good is divisible. Across both experiments, we investigate fairness perceptions in the U.S. population.

\section{Study 1 (No Sensitive Information)}

In this study, our motivation is to investigate how information on an individual task-specific feature (i.e., the candidates' loan repayment rate) influences perceptions of fairness. We present participants with a scenario in which two individuals have each applied for a loan. The participants know no personal information about the two individuals except their loan repayment rates. We choose three allocation rules, described in the following paragraphs, that allow us to formulate qualitative judgments regarding the three fairness definitions.

\subsection{Procedure}
We recruited 200 participants from Amazon Mechanical Turk (MTurk) on March 18-19, 2018. The majority of them identified themselves as white (82\%), 8\% as black, 6\% as Asian or Asian-American, 2\% as Hispanic, and the rest with multiple races. The average age was 39.43 (SD = 12.47). Most (91\%) had attended some college, while almost all other participants had a high school degree or GED. (All demographic information was self-reported.) All participants were U.S. residents, and each were paid \$0.20 for participating.

We presented participants with the scenario presented in Figure \ref{fig:fig1} in the appendix.

This experiment employed a between-subjects design with four conditions. We varied the individual candidates' similarity (dissimilarity) in ability to pay back their loan (i.e., their loan repayment rate), as an operationalization of task-specific similarity (dissimilarity) relevant to the three fairness definitions. Participants were randomly shown one of four loan repayment rates: 55\% and 50\% (Treatment 1), 70\% and 40\% (Treatment 2), 90\% and 10\% (Treatment 3), and 100\% and 20\% (Treatment 4). One treatment had a very small difference between the loan repayment rates of the two candidates (Treatment 1). The next treatment had a larger difference between the loan repayment rates (Treatment 2), with the next two treatments (Treatments 3 and 4) having a much larger difference in their loan repayment rates. Each participant was only shown one Treatment. 
 
We held all other information about the two candidates constant. We then presented participants with three possible decisions for how to allocate the money between the two individuals. The order of the three decisions was counterbalanced. 
 
 Each decision was designed to help us to untangle the three fairness definitions.

\noindent
 \textbf{``All A" Decision. Give all the money to the candidate with the higher payback rate.} 
 This decision is allowed in all treatments under meritocratic fairness as defined \citeauthor{joseph2016fairness} (2016), where a worse applicant is never favored over a better one. It would also be allowed under the definition formulated by \citeauthor{dwork2012fairness} (2012), in the more extreme treatments, and even in every treatment in the case that the similarity metric was very discerning. This decision would not be allowed in any treatment under the calibrated fairness definition \cite{liu2017calibrated}.

\noindent
\textbf{``Equal" Decision. Split the money 50/50 between the candidates, giving \$25,000 to each.}
This decision is allowed in all treatments under \citeauthor{dwork2012fairness} (2012) -- treating similar people similarly. Moreover, under their definition, when two individuals are deemed to be similar to each other, then this is the textit{only} allowable decision (in Treatment 1, for example). This decision is also allowed in all the treatments under the meritocratic definition \cite{joseph2016fairness}, as the candidate with the higher loan repayment rate is given at least as much as the other candidate, and, hence, is weakly favored. The decision, however, would not be allowed in any treatment under calibrated fairness \cite{liu2017calibrated}, since the candidates are not being treated in proportion of their quality (loan repayment rate).

\noindent
\textbf{``Ratio" Decision. Split the money between two candidates in proportion of their loan repayment rates.} 
This decision is allowed in all treatments under calibrated fairness, where resources are divided in proportion to the true quality of the candidates. Moreover, this is the only decision allowed under this definition. This decision could also align with the definition proposed by \citeauthor{dwork2012fairness} (2012), but only for suitably defined similarity metrics that allow the distance between decisions implied by the ratio allocation. Finally, this decision would be allowed under meritocratic fairness \cite{joseph2016fairness} for the same reasons as the ``Equal'' decision. Namely, the candidate with the higher loan repayment rate is weakly favored to the other candidate. 

It is important to note that we are testing  human perceptions regarding the outcomes that different fairness definition allow, not the definitions themselves. However, if a certain definition allows multiple decisions, then we would expect these decisions to receive similar support. Where the perception of the fairness of outcomes is inconsistent with the allowable decisions for a rule, this is worthwhile to understand.  

If it is true that participants most prefer the treating similar people similarly definition, one would expect that they would prefer the ``Equal'' decision to the other two decisions for a wider range of similarity metrics and treatments. If it is true that participants most prefer the meritocratic definition, one would expect no significant difference in support for the three different decisions. If it is true that participants most prefer the calibrated fairness definition, one would expect that the ``Ratio'' decision is perceived as more fair than the other two decisions.

We formulated the following set of hypotheses: 

\noindent
\textbf{Hypothesis 1A.} Across all treatments, participants perceive the ``Ratio" decision as more fair than the ``Equal" decision.

\noindent
\textbf{Hypothesis 1B.} Across all treatments, participants perceive the ``Ratio" decision as more fair than the ``All A" decision.

\noindent
Furthermore, we made the following predictions:

\noindent
\textbf{Hypothesis 2.} Participants perceive the ``Equal" decision as more fair than the ``All A" decision in Treatment 1. That is, participants may view the candidates in Treatment 1 as ``similar enough" to be treated similarly.

\noindent
\textbf{Hypothesis 3.} Participants perceive the ``All A" decision as more fair than the ``Equal" decision in Treatments 3 and 4.

 \begin{figure}[H]
\includegraphics[width=\linewidth]{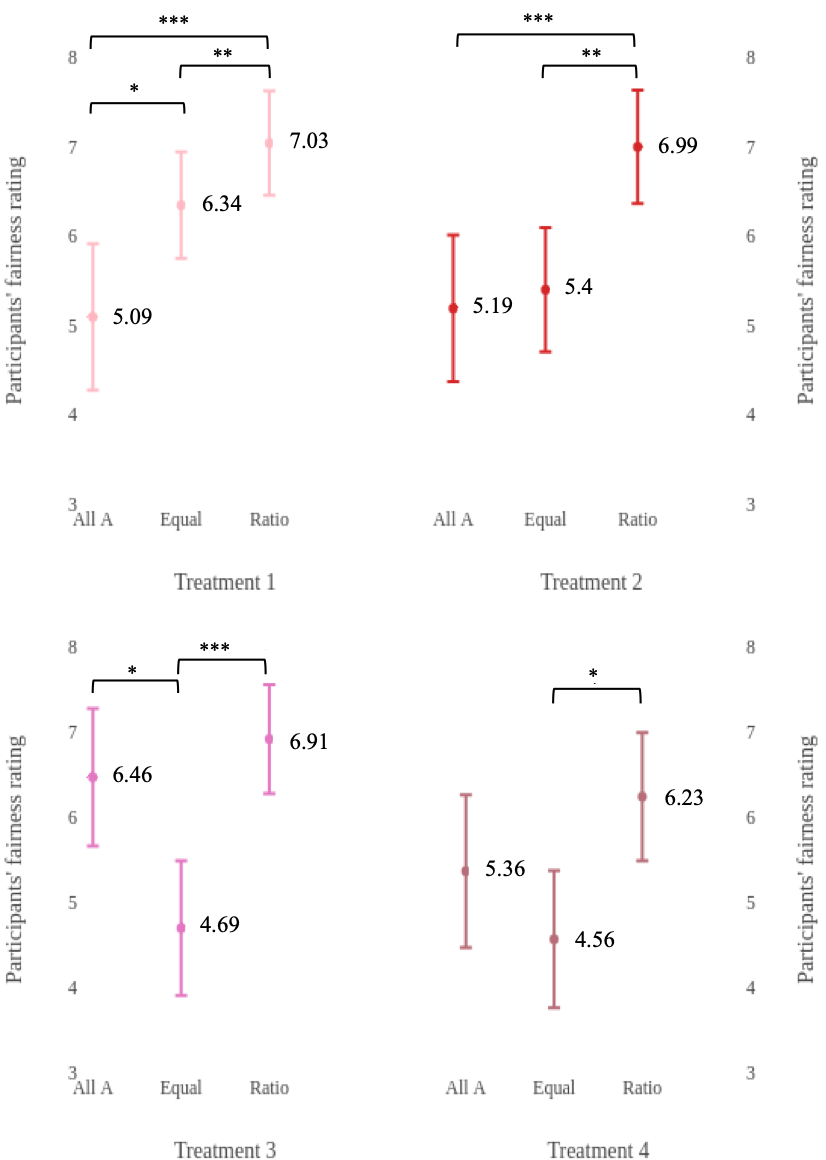}
 \caption{Comparison of means (with 95\% CI) for Study 1.
Where * signifies p \textless 0.05, ** p \textless 0.01, and *** p \textless 0.001.}
 \label{fig:figure2}
 \end{figure} 
 
\subsection{Results}
First, we tested hypotheses H1A and H1B, which conjecture that participants will consider the ``Ratio'' decision as the most fair. We found evidence in support of H1A in all treatments: participants consistently rated dividing the \$50,000 between the two individuals in proportion of their loan repayment rates (the ``Ratio'' decision) as more fair than splitting the \$50,000 equally (the ``Equal'' decision) (see Figure ~\ref{fig:figure2}). We found partial support for H1B: participants rated the ``Ratio'' decision as more fair than the ``All A'' decision in Treatments 1 and 2 (see Figure ~\ref{fig:figure2}).

Second, we found that participants in Treatment 1 rated the ``Equal'' decision as more fair than the ``All A'' definition (see Figure ~\ref{fig:figure2}), supporting H2. We see that when the difference in the loan repayment rates of the individuals was small (5\%), participants perceived the decision to divide the money equally between the individuals as more fair than giving all the money to the individual with the higher loan repayment rate.

Third, we found that participants rated the ``All A'' decision as more fair than the ``Equal'' decision in Treatment 3, but not in Treatment 4 (see Figure ~\ref{fig:figure2}).

\subsection{Discussion}

We see that across all treatments, participants in Study 1 perceived the ``Ratio'' decision -- the only decision that aligns with calibrated fairness -- to be more fair than  the ``Equal'' decision -- the only decision that is always aligned with the treating people similarly definition. One possible explanation is that calibrated fairness implies treating people similarly for a similarity metric \cite{liu2017calibrated} that is based on a notion of merit. 

In Treatments 1 and 2, participants rated the ``Ratio'' decision -- the only decision that aligns with calibrated fairness -- to be more fair than the ``All A'' decision. Note that the meritocratic definition is the only definition that always allows the ``All A'' decision. No significant difference was discovered for Treatments 3 and 4, where one candidate has a much higher repayment rate. 

Furthermore, participants viewed individuals to be similar enough to be treated similarly only when the difference in the applicants' loan repayment rates was very small (approximately 5\%).


 \section{Study 2 (With Sensitive Information)}
In this study, our motivation is to investigate how the addition of sensitive information to information on an individual task-specific feature (i.e., the candidates' loan repayment rate) influences perceptions of fairness.

We employed the same experimental paradigm as in Study 1, presenting participants with the scenario of two individuals applying for a loan, and three possible ways of allocating the loan money. Importantly, in Study 2, in addition to providing information on the individuals' loan repayment rates, we also provided information on the individuals' race. We investigate how information on the candidates' loan repayment rates and the candidates' race influence people's fairness judgments of the three allocation decisions.

\subsection{Procedure}

We recruited a separate sample of 1800 participants from Amazon Mechanical Turk (MTurk) on April 20-21, 2018, none of whom had taken part in Study 1. Most of them identified as white (74\%), 9\% as black, 7\% as Asian or Asian-American, 5\% as Hispanic, and the rest with multiple races. The average age was 36.97 (SD = 12.54). Most (89\%) had attended some college, while almost all other participants had a high school degree or GED. All participants were U.S. residents, and each was paid \$0.20 for participating. (All demographic information was self-reported.)  

We presented participants with the same scenario as in Study 1, but this time also providing the candidates' race and gender. We held the gender of the candidates constant (both were male), and randomized race (black or white). Thus, either the white candidate had the higher loan repayment rate, or the black candidate had the higher loan repayment rate. The question presented to the participants in Study 2 can be found in Figure Figure~\ref{fig:figure5} in the appendix.

We presented the same question and choices for loan allocations, and tested the same hypotheses, as in Study 1.

 \begin{figure}[H]
\includegraphics[width=\linewidth]{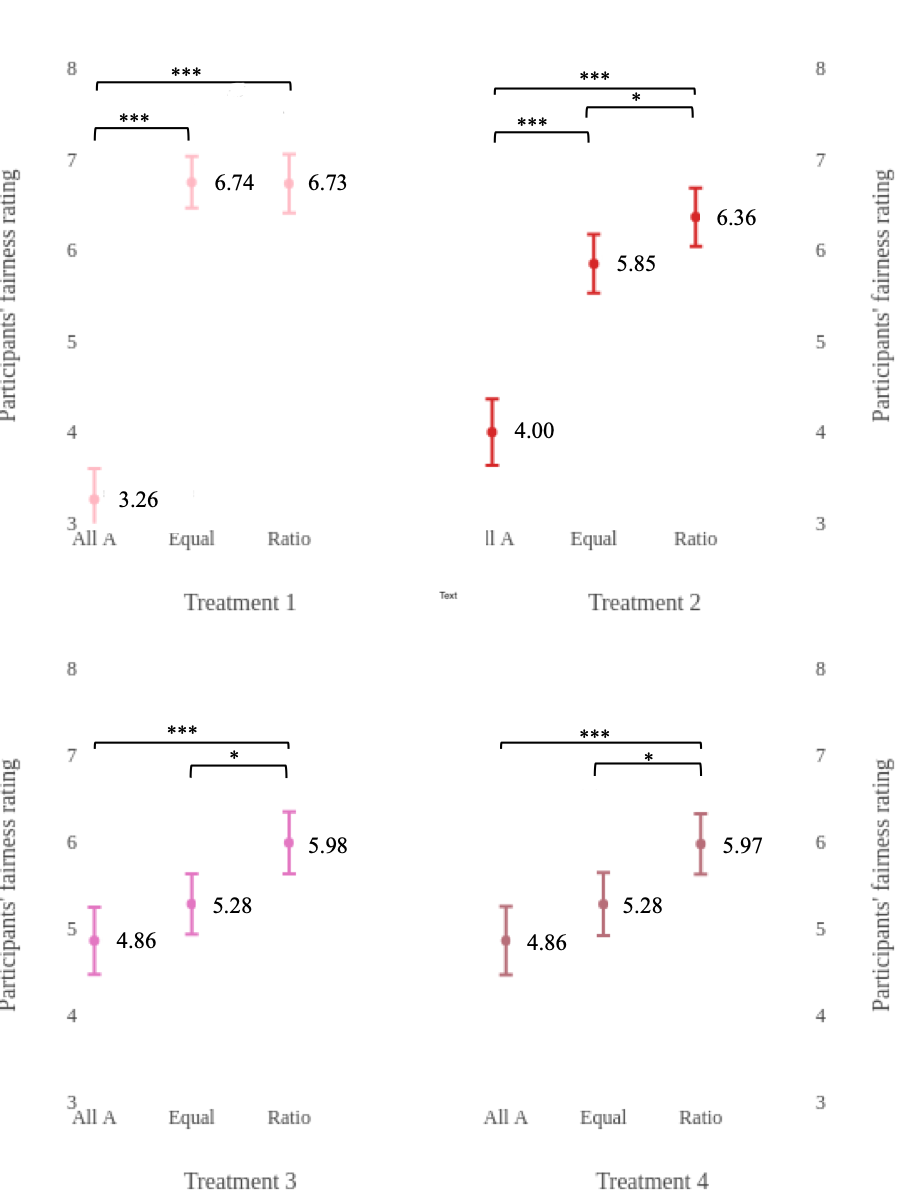}
 \caption{Comparison of means (with 95\% CI) for Study 2 (when the individual with the higher loan repayment rate is white).
Where * signifies p \textless 0.05, ** p \textless 0.01, and *** p \textless 0.001.}
 \label{fig:figure3}
 \end{figure}
 
\subsection{Results}
\begin{figure}[H]
\includegraphics[width=\linewidth]{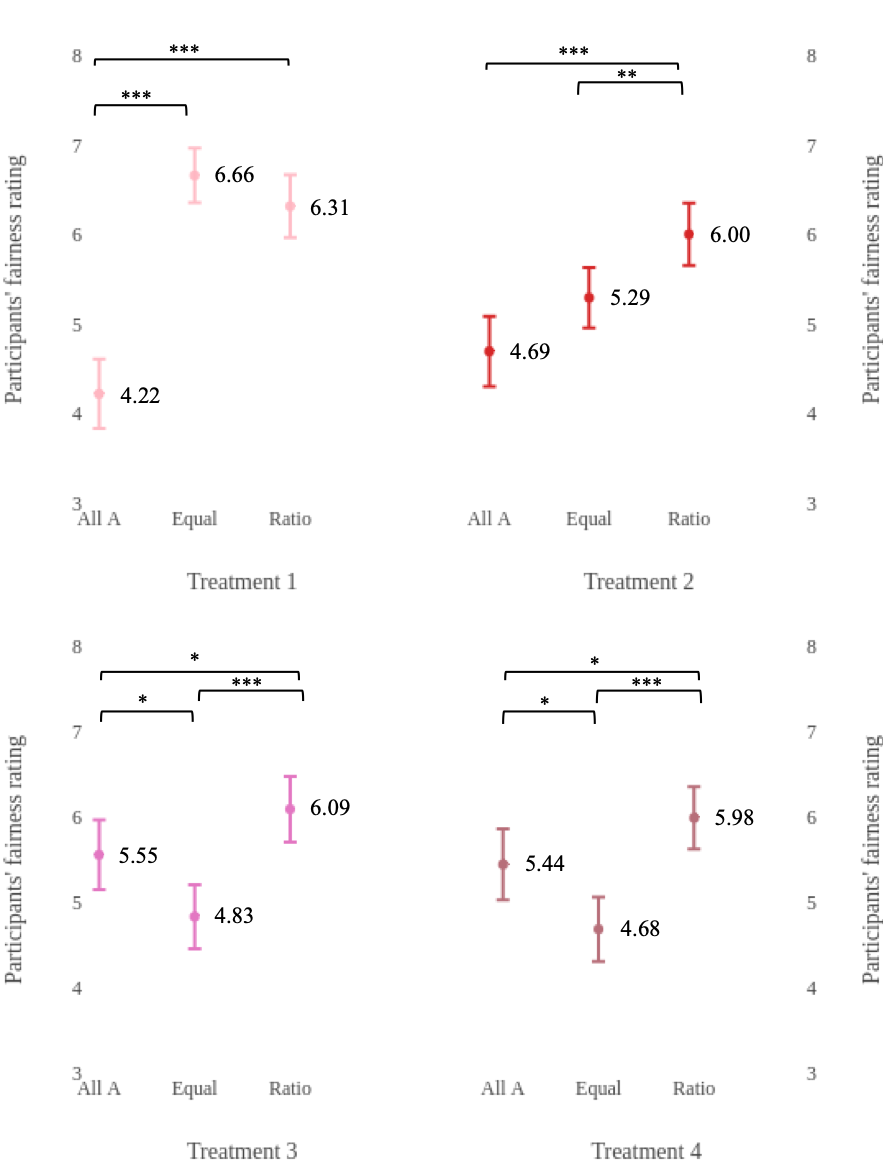}
 \caption{Comparison of means (with 95\% CI) for Study 2 (when the individual with the higher loan repayment rate is black).
Where * signifies p \textless 0.05, ** p \textless 0.01, and *** p \textless 0.001.}
 \label{fig:figure4}
 \end{figure}

We found that participants viewed the ``Ratio'' decision as more fair than the ``Equal'' decision in Treatments 2, 3, and 4, regardless of race, in support of H1A. We also found that participants viewed the ``Ratio'' decision as more fair than the ``All A'' decision in all treatments, regardless of race, thus supporting H1B. (See Figures ~\ref{fig:figure3} and ~\ref{fig:figure4}.) Thus, participants in Study 2 consistently gave most support to the decision to divide the \$50,000 between the two individuals in proportion to their loan repayment rates.

Furthermore, we found that participants viewed the ``Equal'' decision as more fair than the ``All A'' decision in Treatment 1, regardless of race, in support of H2 (see Figures ~\ref{fig:figure3} and ~\ref{fig:figure4}). Participants also rated the ``Equal'' decision as more fair than the ``All A'' decision in Treatment 2, but only when the candidate with the higher repayment rate was white (see Figure ~\ref{fig:figure3}).

When the difference between the two candidates' repayment rates was larger (Treatments 3 and 4), participants viewed the ``All A'' decision as more fair than the ``Equal'' decision but only when the candidate with the higher repayment rate was black (see Figure ~\ref{fig:figure4}). By contrast, when the candidate with the higher loan repayment rate was white, participants did not rate the two decisions differently (see Figure ~\ref{fig:figure3}).

 \subsection{Discussion} 
 In Study 2,  we tested whether participants' perceptions of these three fairness definitions could be influenced by additional information regarding the candidates' race. 
 
Our results show that participants perceived the ``Ratio'' decision to be more fair than the other two, hence supporting the results from Study 1 and the related discussion. These results are not dependent on the race attribute. Furthermore, regardless of race, when the difference between the  loan repayment rates was small (Treatment 1), participants preferred the ``Equal'' decision to the ``All A'' decision. This supports the corresponding results from Study 1, Treatment 1, which indicate that one should account for similarity of individuals when designing fair rules.   
 
However, we also found evidence that race does affect participants' perception of fairness. When the difference in loan repayment rates was larger (Treatments 3 and 4), participants rated the ``All A'' decision as more fair than the ``Equal'' decision, but only when the candidate with the higher repayment rate was black. These results suggest a boundary condition of H3: people may support giving all the loan money to the candidate with the higher payback rate, compared to splitting the money equally, when the candidate with the higher payback rate is a member of a group that is historically disadvantaged. 


Each definition, from meritocratic to similarity to calibrated fairness is successively stronger in our context, ruling out additional decisions. In this light, it is interesting that the ratio decision is most preferred, providing support for the calibrated fairness definition, even though this definition is the strongest of the three in the present context. When historically disadvantaged individuals have a higher repayment rate, participants are more supportive of more decisive allocations in favor of the stronger, and historically disadvantaged, individual.


\section{Conclusion} 

People broadly show a preference for the ``Ratio'' decision, which is
indicative of their support for the calibrated fairness definition
\cite{liu2017calibrated}, as compared to the treating similar people
similarly \cite{dwork2012fairness} and meritocratic
\citeauthor{joseph2016fairness} definitions. We also find in Study 2
some support for the principle of affirmative action.


Through the use of crowdsourcing, we can elicit information on public
attitudes towards different definitions of algorithmic fairness, and
how individual characteristics, such as task-specific features (e.g.,
loan repayment rates) and sensitive attributes (e.g., race) could be
relevant in fair decision-making. Understanding public attitudes can
help to continue a dialogue between technologists and ethicists in the
design of algorithms that make decisions of consequence to the public.
For example, the three fairness definitions examined here agree that,
conditioned on the task-specific metric, an attribute such as race
should not be relevant to decision-making. Yet, we find some
treatments under which people's attitudes about loan decisions change
when race is provided to the context.

This paper opens up several directions for future research. Beyond testing additional definitions, future experiments could in addition specify whether the decision was made by a human or an algorithm. Psychological theories of mind may influence people's fairness judgments. Third, future work could investigate how people perceive fairness in other contexts, such as university admissions or bail decisions, where there is no divisible resource but rather a definite decision needs to be made, and in the university case in the context of a resource constraint. Fourth, further research could examine why the availability of additional personal or sensitive information influences perceptions of fairness. Why do people consider factors such as race important for their fairness ratings? And to what extent are people willing to endorse affirmative action in defining algorithmic fairness? Finally, it is important to consider how to incorporate the general public's views into algorithmic decision-making. 

These  results are only the start of a research program on understanding ordinary people's fairness judgments of definitions of algorithmic fairness. As the literature on moral psychology has shown, people often make inconsistent and unreasoned moral judgments \cite{greene2014moral}. Indeed,  research on moral judgments in regard to the decisions made by autonomous vehicles (the ``moral machine'') has shown that people approve of utilitarian autonomous vehicles, but are unwilling to purchase utilitarian autonomous vehicles for themselves \cite{bonnefon2016social}. On the other hand,  research in moral psychology shows that people can engage in sophisticated moral reasoning, thinking in an impartial, bias-free way, resulting in moral judgments that favor the greater good \cite{huang2019veil}. Future research could investigate how moral reasoning interventions could influence people's fairness judgments in the domain of algorithmic fairness.

 \bibliographystyle{aaai}
\bibliography{aies19}

\newpage

\newpage
\appendix
\section{Appendix}

To be eligible to take our surveys, the Amazon Mechanical Turk workers had to be located in the United States of America. We stipulated this restriction via TurkPrime, which is a platform for performing crowdsourced research when using Amazon Mechanical Turk.

Amazon Mechanical Turk workers ('MTurker') could only participate in on one of the two studies, and not both.

The first section contains all questions the workers were asked in the studies. The second section contains the demographics data for the respondents of both studies. The third section details how many participants in each study were shown the four treatments.

\subsection{Questions from the studies}
\begin{figure}[H]
\includegraphics[width=1.5\linewidth]{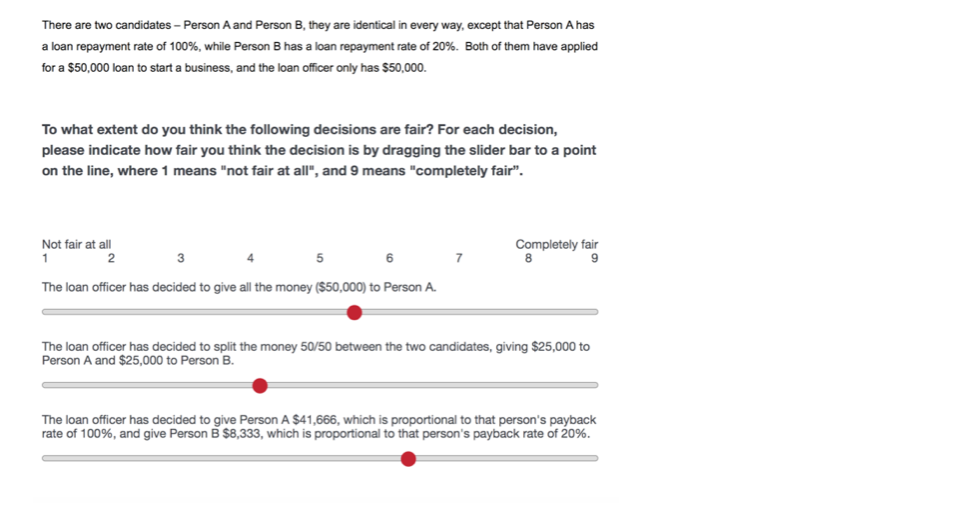}
 \caption{Question presented to the participants in Study 1.}
 \label{fig:fig1}
 \end{figure}

\begin{figure}[H]
\includegraphics[width=\linewidth]{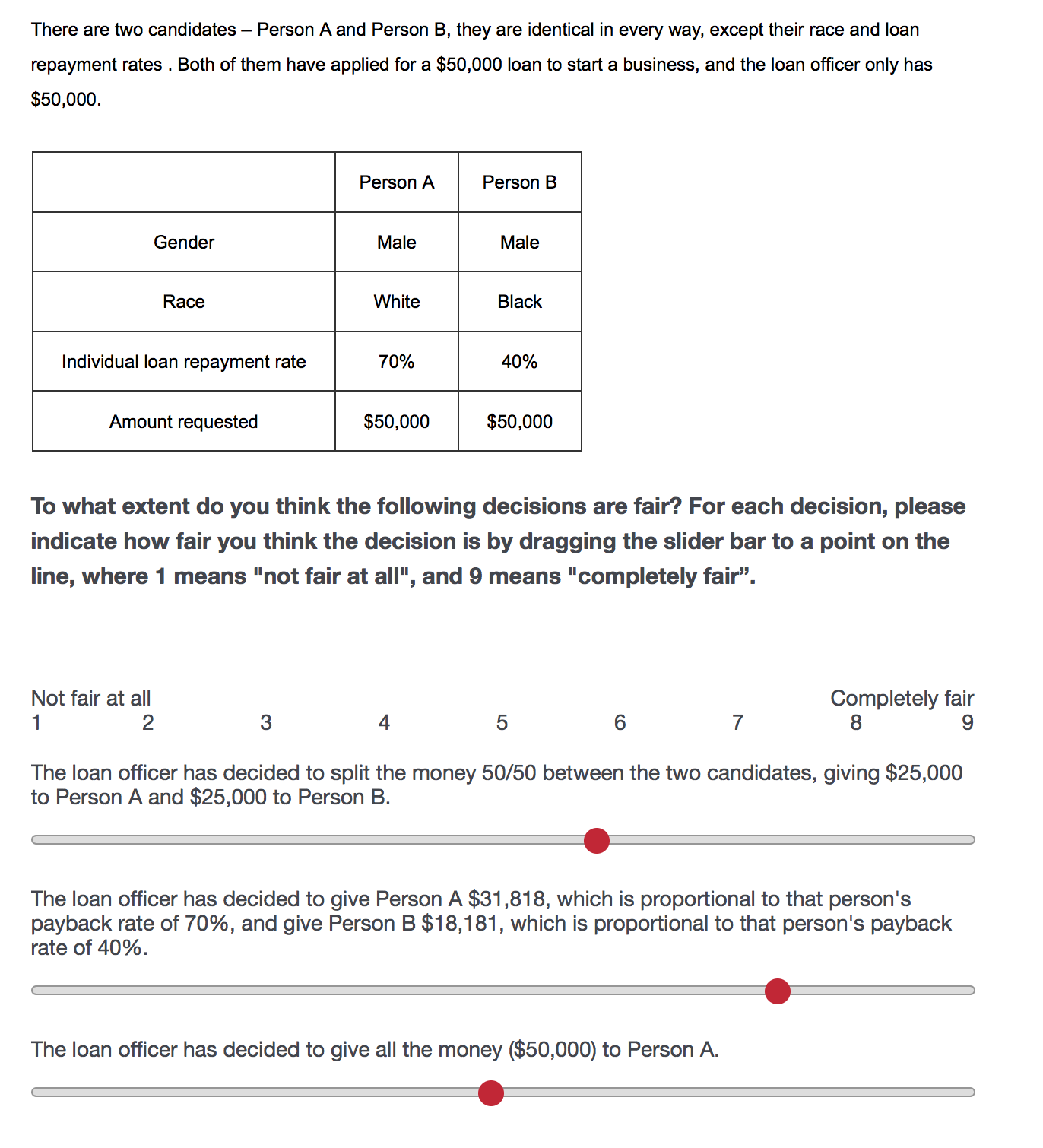}
 \caption{Question presented to the participants in Study 2.}
 \label{fig:figure5}
 \end{figure}

The question asked in Study 1 is presented in Figure ~\ref{fig:fig1}. The question asked in Study 2 is presented in Figure ~\ref{fig:figure5}. 

\subsubsection{Demographics questions}
Every Amazon Mechanical Turk worker in both the surveys was asked that study's question. After they answered that, they were asked the following demographics questions that were voluntary to answer. Since they were voluntary to answer, not all respondents answered them, although most (\textgreater 90\%) answered some or all of them.

\begin{enumerate}

\item What state do you live in?
\item Do you identify as:
\begin{itemize}
\item[$\ocircle$] Male
\item[$\ocircle$] Female
\item[$\ocircle$] Other (please specify): \_\_\_\_\_\_\_\_\_

\end{itemize}

\item What is the highest level of school you have completed or the highest degree you have received?
\begin{itemize}
\item[$\ocircle$] Less than high school degree
\item[$\ocircle$] High school degree or equivalent
\item[$\ocircle$] Some college but no degree
\item[$\ocircle$] Associate degree
\item[$\ocircle$] Bachelor degree
\item[$\ocircle$] Graduate degree
\end{itemize}

\item Do you identify as:
\begin{itemize}
\item[$\Box$] Spanish, Hispanic, or Latino
\item[$\Box$] White
\item[$\Box$] Black or African-American
\item[$\Box$] American-Indian or Alaskan Native
\item[$\Box$] Asian
\item[$\Box$] Asian-American
\item[$\Box$] Native Hawaiian or other Pacific Islander
\item[$\Box$] Other (please specify): \_\_\_\_\_\_\_\_\_
\end{itemize}

\item In what type of community do you live:
\begin{itemize}
\item[$\Box$] City or urban community
\item[$\Box$] Suburban community
\item[$\Box$] Rural community
\item[$\Box$] Other (please specify): \_\_\_\_\_\_\_\_\_
\end{itemize}

\item What is your age?

\item Which political party do you identify with?
\begin{itemize}
\item[$\Box$] Democratic Party
\item[$\Box$] Republican Party
\item[$\Box$] Green Party
\item[$\Box$] Libertarian Party
\item[$\Box$] Independent
\item[$\Box$] Other (please specify): \_\_\_\_\_\_\_\_\_
\end{itemize}

\end{enumerate}

\subsection{Study 1: Demographic information of the participants}

 \begin{figure}[H]
\includegraphics[width=\linewidth]{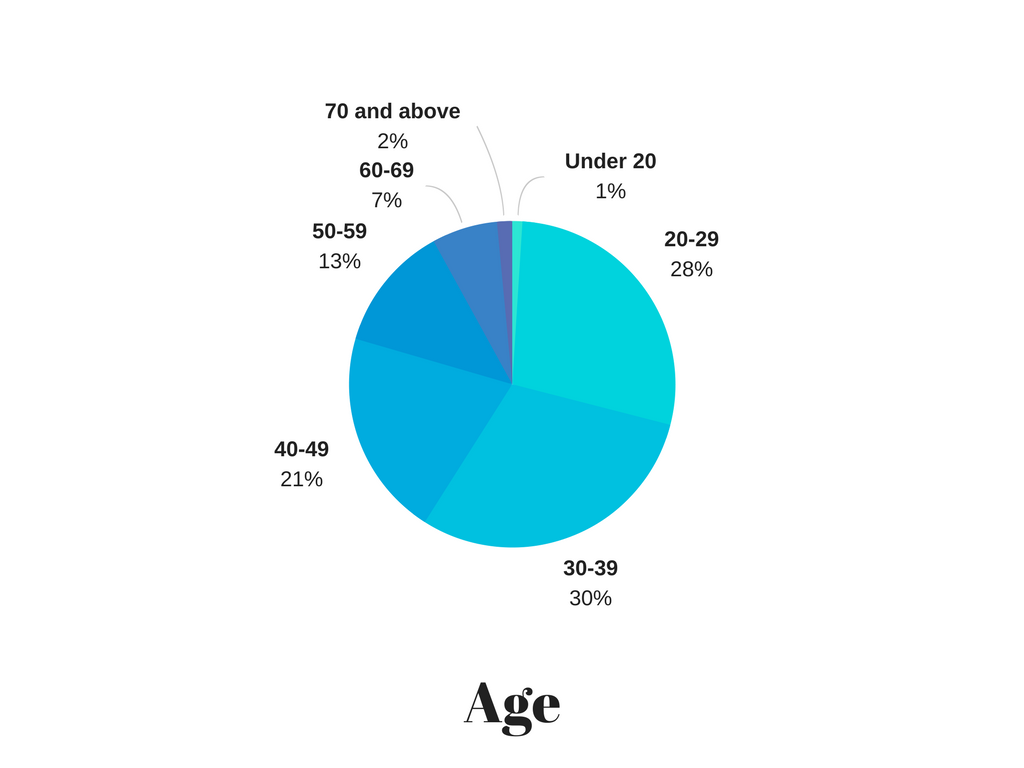}
 \caption{Age distribution of the participants in Study 1.}
 \label{fig:figure6}
 \end{figure}

 \begin{figure}[H]
\includegraphics[width=\linewidth]{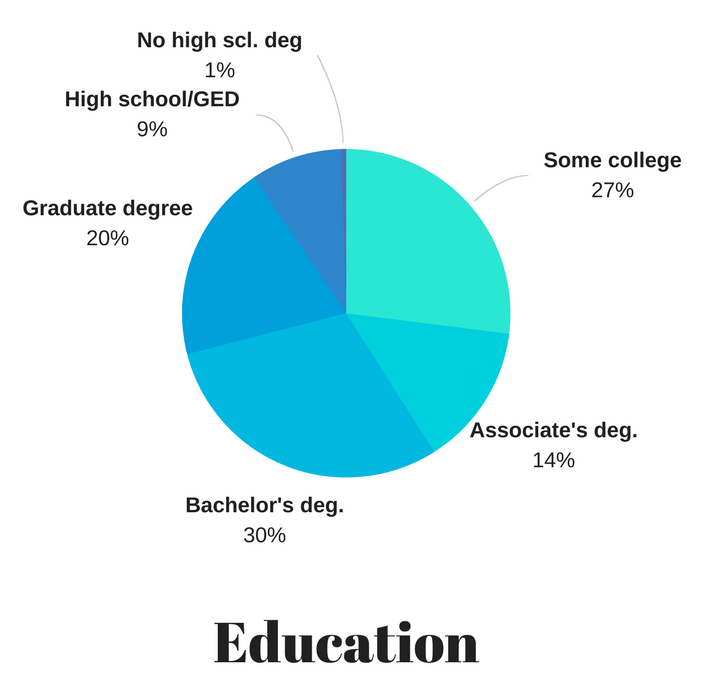}
 \caption{Education of the participants in Study 1.}
 \label{fig:figure7}
 \end{figure}

 \begin{figure}[H]
\includegraphics[width=\linewidth]{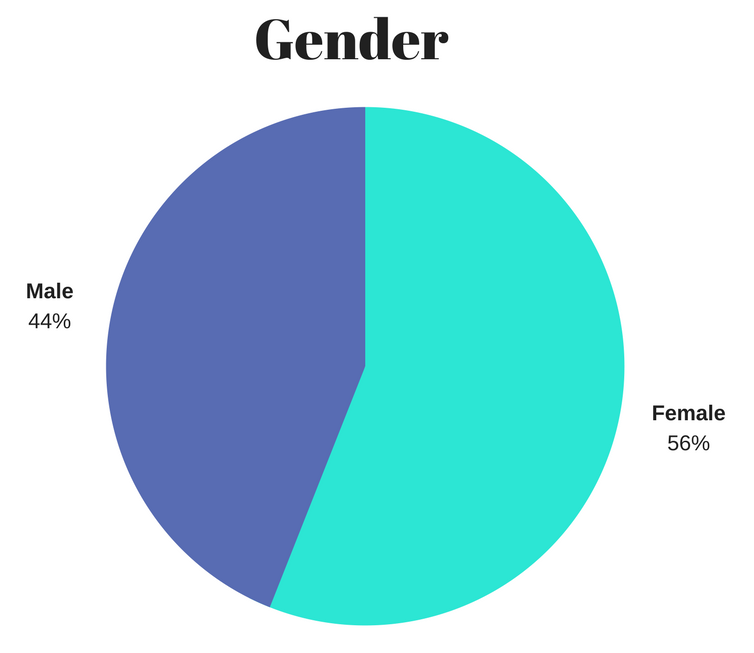}
 \caption{Gender breakdown of the participants in Study 1.}
 \label{fig:figure8}
 \end{figure}

 \begin{figure}[H]
\includegraphics[width=\linewidth]{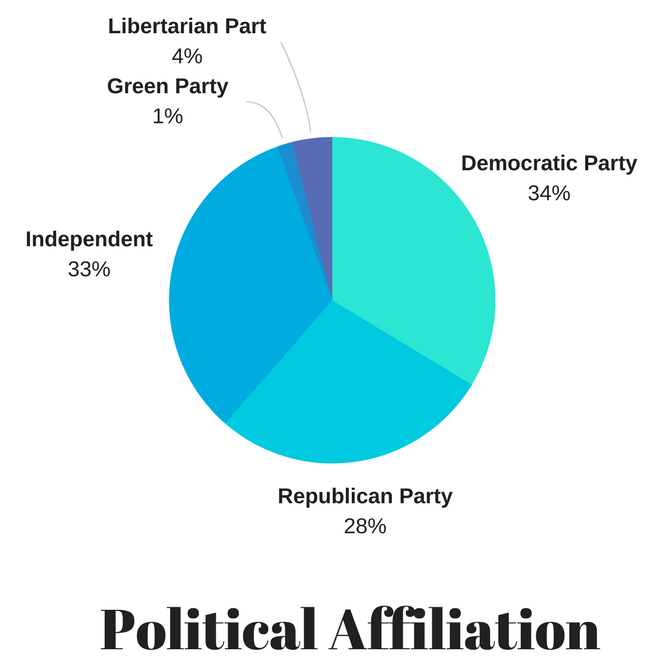}
 \caption{Political Affiliation of the participants in Study 1.}
 \label{fig:figure9}
 \end{figure}
 
  \begin{figure}[H]
\includegraphics[width=\linewidth]{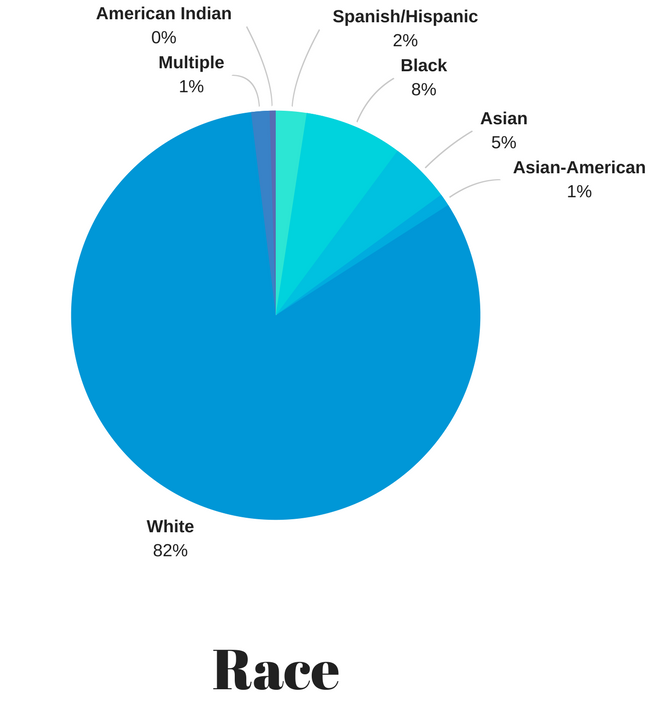}
 \caption{Race of the participants in Study 1.}
 \label{fig:figure10}
 \end{figure}

  \begin{figure}[H]
\includegraphics[width=\linewidth]{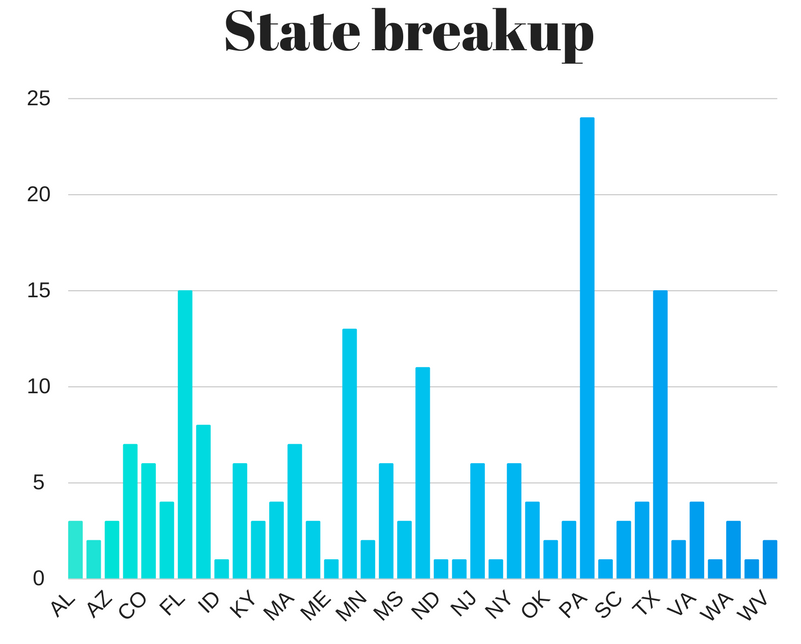}
 \caption{Breakup by state of the participants in Study 1.}
 \label{fig:figure11}
 \end{figure}
 
   \begin{figure}[H]
\includegraphics[width=\linewidth]{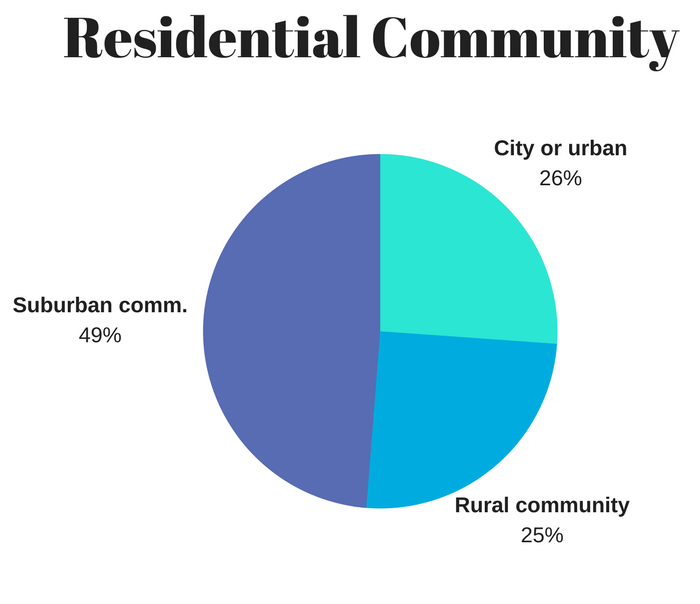}
 \caption{Residential breakdown of the participants in Study 1.}
 \label{fig:figure12}
 \end{figure} 
 
\subsection{Study 2: Demographic information of the participants}
 
  \begin{figure}[H]
\includegraphics[width=\linewidth]{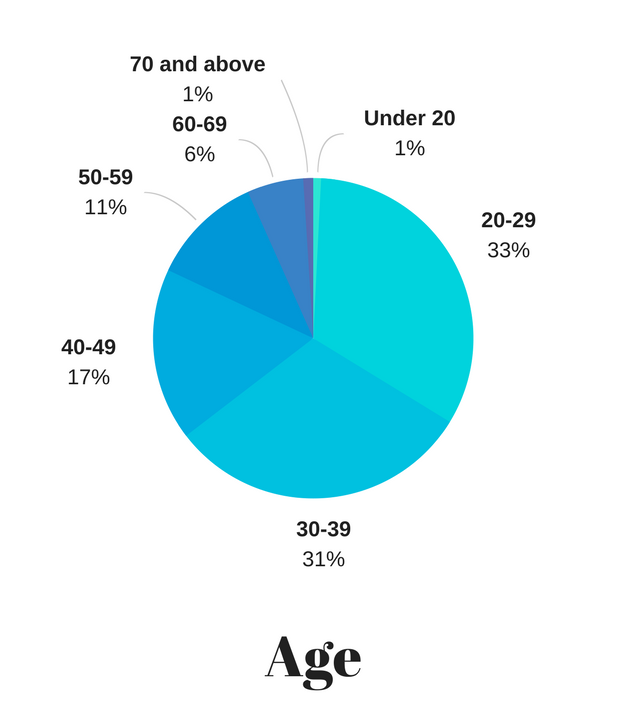}
 \caption{Age distribution of the participants in Study 2.}
 \label{fig:figure13}
 \end{figure}
 
  \begin{figure}[H]
\includegraphics[width=\linewidth]{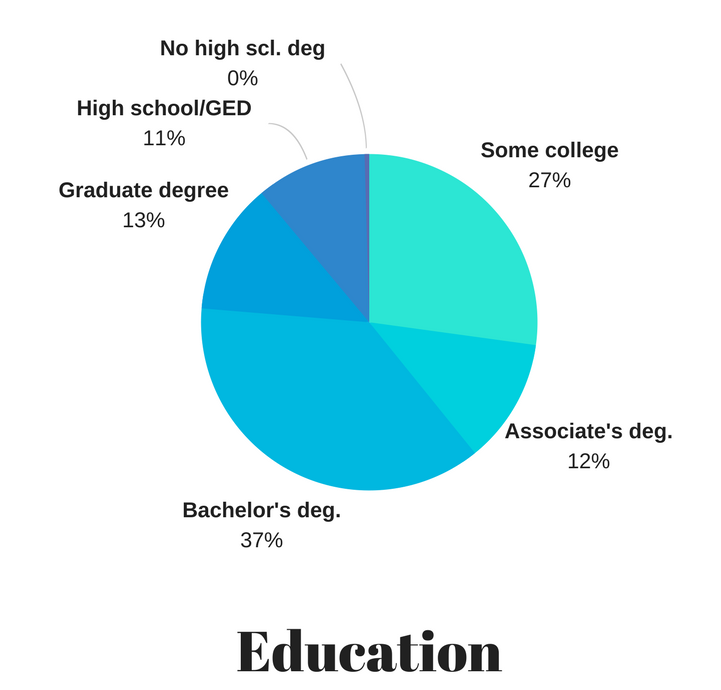}
 \caption{Education of the participants in Study 2.}
 \label{fig:figure14}
 \end{figure}
 
  \begin{figure}[H]
\includegraphics[width=\linewidth]{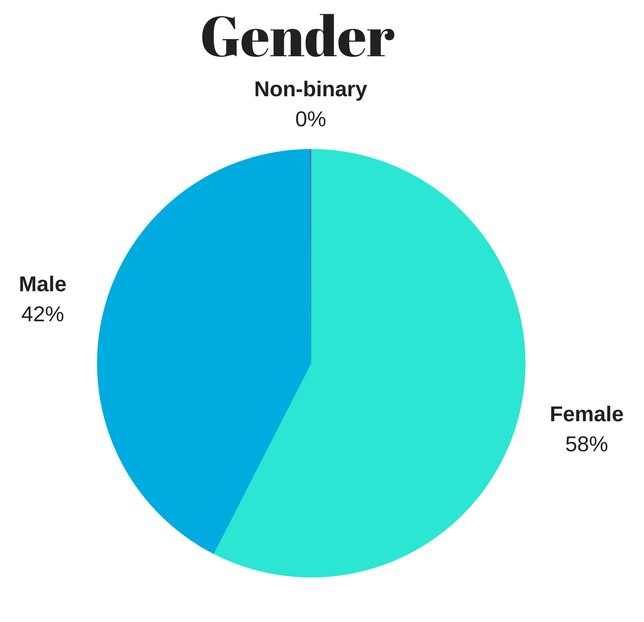}
 \caption{Gender breakdown of the participants in Study 2.}
 \label{fig:figure15}
 \end{figure}
  
  \begin{figure}[H]
\includegraphics[width=\linewidth]{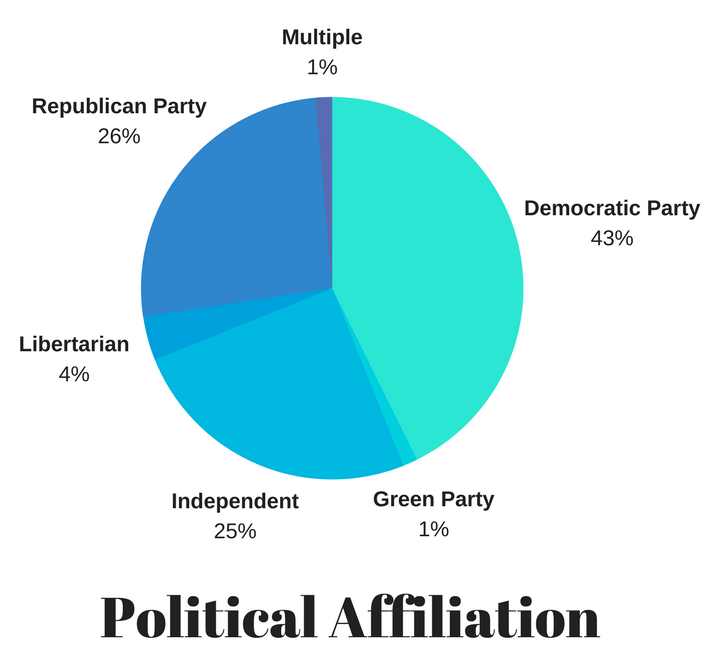}
 \caption{Political Affiliation of the participants in Study 2.}
 \label{fig:figure16}
 \end{figure}

  \begin{figure}[H]
\includegraphics[width=\linewidth]{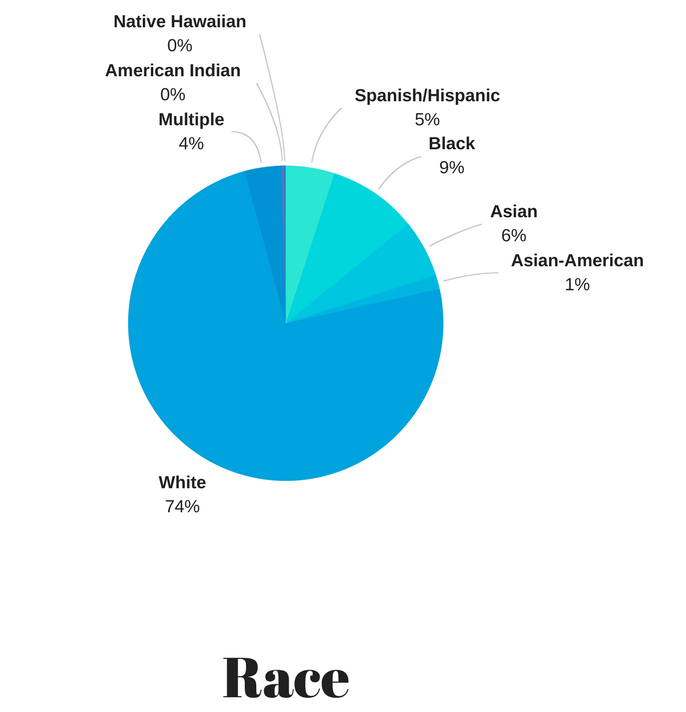}
 \caption{Race of the participants in Study 2.}
 \label{fig:figure17}
 \end{figure}

  \begin{figure}[H]
\includegraphics[width=\linewidth]{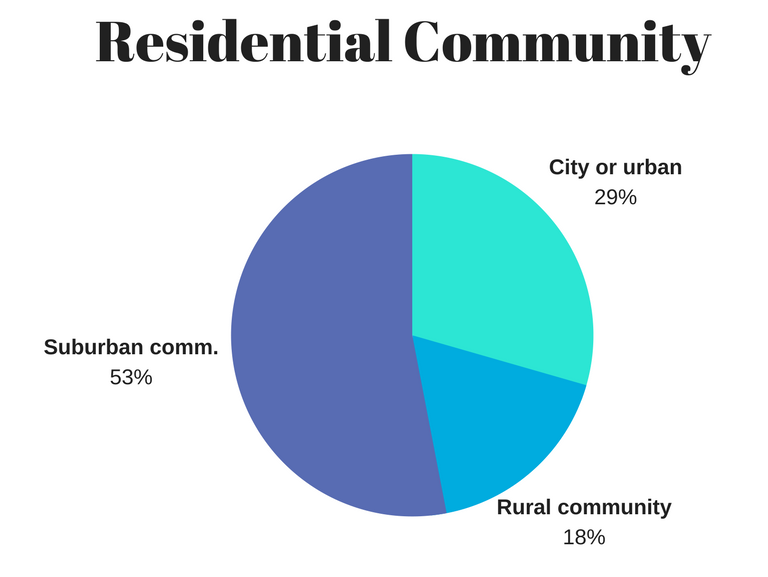}
 \caption{Residential breakdown of the participants in Study 2.}
 \label{fig:figure18}
 \end{figure}

\end{document}